\documentclass[10pt,twocolumn,letterpaper]{article}

\usepackage[pdftex]{graphicx}
\usepackage{booktabs}
\usepackage{iccv}
\usepackage{times}
\usepackage{epsfig}
\usepackage{amsmath}
\usepackage{amssymb}
\usepackage{pifont}
\newcommand{\cmark}{\ding{51}}%
\usepackage[breaklinks=true,bookmarks=false]{hyperref}
\hypersetup{colorlinks,urlcolor=}

\iccvfinalcopy 

\def\iccvPaperID{1303} 

\ificcvfinal\fi

\begin{document}

\renewcommand{\thefootnote}{\fnsymbol{footnote}}
\title{\vspace{-2mm}Joint Detection and Recounting of Abnormal Events \\ by Learning Deep Generic Knowledge$^*$\vspace{-2mm}}

\author{Ryota Hinami$^{1, 2}$, Tao Mei$^3$, and Shin'ichi Satoh$^{2, 1}$\\
$^{1}$The University of Tokyo, 
$^{2}$National Institute of Infomatics, $^{3}$Microsoft Research Asia \\
{\tt\small hinami@nii.ac.jp, tmei@microsoft.com, satoh@nii.ac.jp}
}


\maketitle
\thispagestyle{empty}

\begin{abstract}
This paper addresses the problem of joint detection and recounting of abnormal events in videos.
Recounting of abnormal events, i.e., explaining why they are judged to be abnormal, 
is an unexplored but critical task in video surveillance, 
because it helps human observers quickly judge if they are false alarms or not.
To describe the events in the human-understandable form for event recounting,
learning generic knowledge about visual concepts (e.g., object and action) is crucial.
Although convolutional neural networks (CNNs) have achieved promising results in learning such concepts, 
it remains an open question as to how to effectively use CNNs for abnormal event detection, 
mainly due to the environment-dependent nature of the anomaly detection.
In this paper, we tackle this problem by integrating a generic CNN model and
environment-dependent anomaly detectors.
Our approach first learns CNN with multiple visual tasks
to exploit semantic information that is useful 
for detecting and recounting abnormal events.
By appropriately plugging the model into anomaly detectors,
we can detect and recount abnormal events while taking advantage of the discriminative power of CNNs.
Our approach outperforms the state-of-the-art on Avenue and UCSD Ped2 
benchmarks for abnormal event detection 
and also produces promising results of abnormal event recounting.

\end{abstract}

\footnotetext[1]{This work was conducted when the first author was a research intern at Microsoft Research Asia.}
\renewcommand{\thefootnote}{\arabic{footnote}}
\setcounter{footnote}{0}

\section{Introduction} 
\begin{figure}[t] 
\begin{center}
\includegraphics[width=1.00\linewidth]{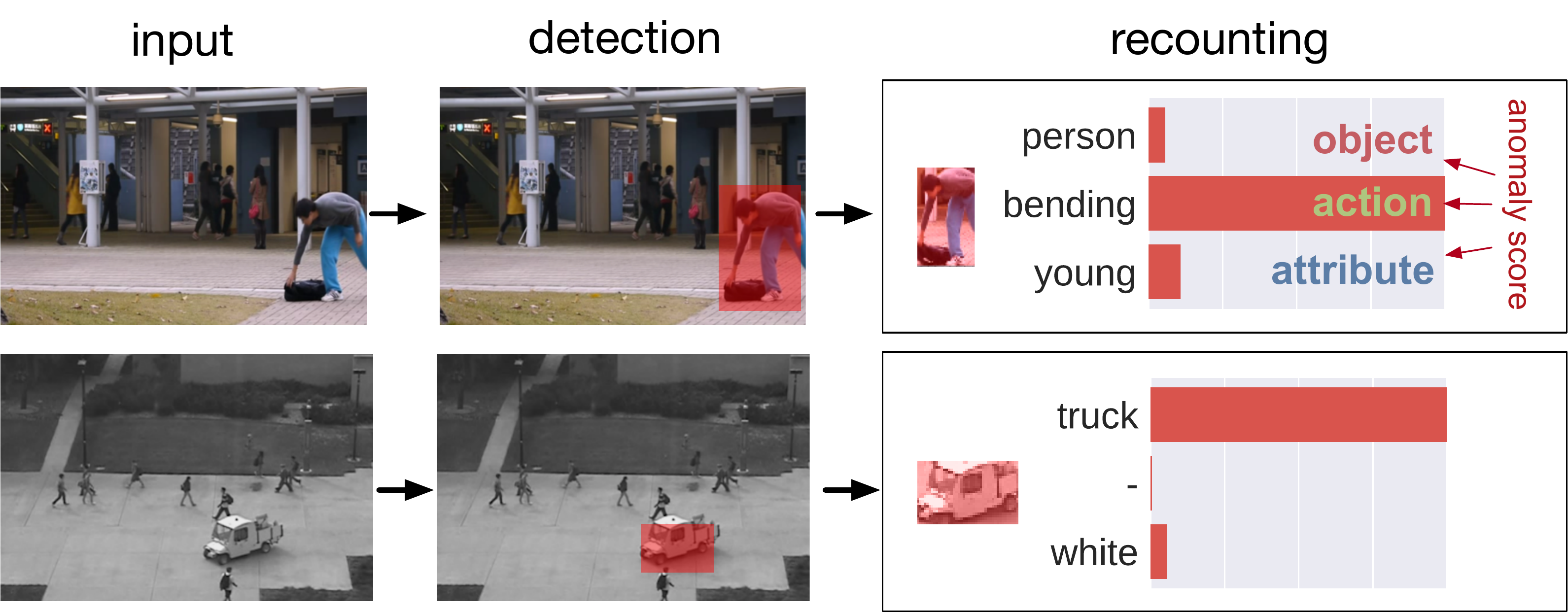}
\caption{Our approach detects abnormal events and also
recounts why they were judged to be abnormal by 
predicting visual concepts and anomaly scores of each concept.}
\label{fig:top} 
\vspace{-4mm}
\end{center} 
\end{figure}

Detecting abnormal events in videos is crucial for video surveillance.
While automatic anomaly detection can free people from having to monitor videos,
we still have to check videos when the systems raise alerts, 
and this still involves immense costs.
If systems can explain what is happening and 
assess why events are abnormal,
we can quickly identify unimportant events without having to check videos.
Such processes that explain the evidence in detecting events is called {\it event recounting},
which was attempted as a multimedia event recounting (MER) task in TRECVid 
\footnote{http://www.nist.gov/itl/iad/mig/mer12.cfm}
but has not been explored in the field of abnormal event detection.
Recounting abnormal events is also useful in understanding the behavior of algorithms.
Analyzing the evidence of detecting abnormal events should disclose
potential problems with current algorithms and indicate possible future directions.
The main goal of the research presented in this paper was to develop a framework 
that could jointly detect and recount abnormal events, as shown in Fig.~\ref{fig:top}. 

Abnormal events are generally defined as irregular events that deviate from normal ones.
Since normal behavior differs according to the environment, 
the target of detection in abnormal event detection depends on the environment
(e.g., `riding a bike' is abnormal indoors while it is normal on cycling roads). 
In other words, {\it positive} in anomaly detection has the nature of {\it environment-dependent}, 
wherein only negative samples are given as training data and positive in the environment 
is defined by these negative samples.
This is different from most other computer vision tasks 
(e.g., `pedestrian' is always positive on a pedestrian detection task).
Since positive samples are not given in anomaly detection, 
detectors of abnormal events cannot be learned in a supervised way. 
Instead, the standard approach to anomaly detection is 
1) learning an environment-dependent normal model using training samples, 
and 2) detecting outliers from the learned model.

However, 
learning knowledge about basic visual concepts is essential for event {\it recounting}.
The event in the example in Fig.~\ref{fig:top} is explained as 
`person', `bending', and `young', because it has knowledge of these concepts.
We call such knowledge {\it generic knowledge}.
We consider generic knowledge to be essential for recounting 
and also to contribute to accurately detecting abnormal events.
Since people also detect anomalies after recognizing the objects and actions,
employing generic knowledge in abnormal event detection fits in well with our intuition.

Convolutional neural networks (CNNs) have proven successful 
in learning visual concepts such as object categories and actions.
CNNs classify or detect target concepts with high degrees 
of accuracy by learning them with numerous positive samples.
However, positive samples are not given in anomaly detection
due to its environment-dependent nature.
This is the main reason that CNNs still have not been successful in anomaly detection and
most approaches still rely on low-level hand-crafted features.
If we can fully exploit the representation power of CNNs, the 
performance of anomaly detection will be significantly improved as it is in other tasks.
Moreover, its learned generic knowledge will help to recount abnormal events.

This paper presents a framework that jointly detects and recounts abnormal events by 
integrating generic and environment-specific knowledge into a unified framework.
A model based on Fast R-CNN~\cite{Girshick2015} 
is trained on large supervised datasets to learn generic knowledge.
Multi-task learning is incorporated into Fast R-CNN to learn three types of concepts, 
actions, objects, and attributes, in one model.
Then, environment-specific knowledge is learned using anomaly detectors.
Unlike previous approaches that have trained anomaly detectors on low-level features,
our anomaly detector is trained on more semantic spaces by using CNN outputs (i.e., 
deep features and classification scores) as features.
Our main contributions are:
\begin{itemize}
\setlength{\leftskip}{-7pt}
\setlength{\itemsep}{-4pt}
\setlength{\parsep}{0pt}
\item{We address a new problem, i.e., joint abnormal event detection and recounting,
which is important for practical surveillance applications as well as 
understanding the behavior of the abnormal event detection algorithm.}
\item{We incorporate the learning of basic visual concepts 
into the abnormal event detection framework.
Our concept-aware model opens up interesting directions
for higher-level abnormal event detection.
}
\item{
Our approach based on multi-task Fast R-CNN 
achieves superior performance over other methods on several benchmarks
and demonstrates the effectiveness of deep CNN features in abnormal event detection.
}
\end{itemize}


\begin{figure*}[t] 
\begin{center}
\includegraphics[width=1.00\linewidth]{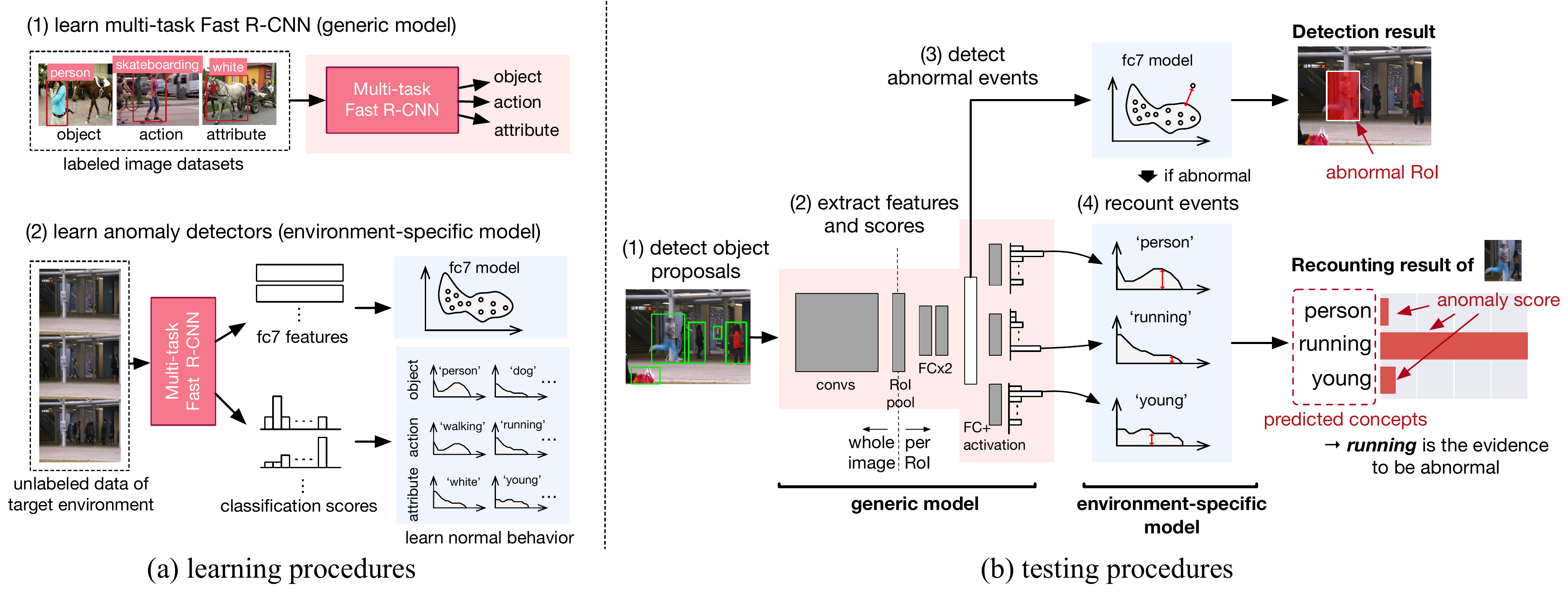}
\vspace{-5mm}
\caption{Overview of our approach. (a) illustrates our learning procedures of two types of models: generic and environment-specific models, and 
(b) shows our testing procedure of joint detection and recounting abnormal events.}
\vspace{-7mm}
\label{fig:overview} 
\end{center} 
\end{figure*}

\section{Related Work} 
The approach of anomaly detection first involves modeling normal behavior 
and then detecting samples that deviate from it.
Modeling normal patterns of object trajectories is one standard 
approach~\cite{Basharat2008,Jiang2011,Siebel2002,Zhang2009} 
to anomaly detection in videos.
While it can capture long-term object-level semantics, 
tracking fails in crowded or cluttered scenes.
An alternative approach is modeling appearance and activity patterns
using low-level features extracted from local regions, 
which is a current standard approach, especially in crowded scenes.
This approach can be divided into two stages:
{\it local anomaly detection} assigning anomaly score to each local region independently, and
{\it globally consistent inference} integrating local anomaly scores into a globally consistent anomaly map with statistical inferences.

Local anomaly detection can be seen as a simple novelty detection problem~\cite{Pimentel2014}: 
a model of normality is inferred using a set of normal features ${\bf X}$ as training data,
and used to assign novelty scores (anomaly scores) $z(x)$ to test sample $x$.
Novelty detectors used in video anomaly detection include
distance-based~\cite{Saligrama2012},
reconstruction-based (e.g., autoencoders~\cite{Hasan2016,Sabokrou2015},
sparse coding~\cite{Cong2011,Lu2013,Zhao2011}), 
domain-based (one-class SVM~\cite{Xu2015}),
and probabilistic methods (e.g.,
mixture of dynamic texture~\cite{Mahadevan2010}, 
mixture of probabilistic PCA~\cite{Kim2009}), 
following the categories in the review by Pimentel et al.~\cite{Pimentel2014}.
These models are generally built 
on the low-level features (e.g., HOG and HOF) 
extracted from densely sampled local patches.
Several recent approaches have investigated the learning-based 
features using autoencoders~\cite{Sabokrou2015,Xu2015},
which minimize reconstruction errors without using any labeled 
data for training.
Antic and Ommer~\cite{Antic2011} detected object hypotheses by video parsing instead of dense sampling,
although they relied on background subtraction that was not robust to illumination changes.

Globally consistent inference was introduced in several approaches 
to guarantee the consistency of local anomaly scores.
Kratz et al.~\cite{Kratz2009} enforce temporal consistency
by modeling temporal sequences with hidden Markov models (HMM).
The spatial consistency is also introduced in 
several studies using Markov random field 
(MRF)~\cite{Benezeth2009,Kim2009,Li2014}
to capture spatial interdependencies between local regions.

While recent approaches have placed emphasis on global consistency~\cite{Cheng2015,DelGiorno2016,Li2014}, 
it is defined on top of the local anomaly scores as explained above.  
Besides, several critical issues remain in local anomaly detection.  
Despite the great success of CNN approaches in many visual tasks, 
the application of CNN to abnormal event detection is yet to be explored.  
Normally CNN requires rich supervised information 
(positive/negative, ranking, etc.) and abundant training data. 
However, supervised information is unavailable for anomaly detection 
by definition.
Hasan et al.~\cite{Hasan2016} learned temporal regularity in videos 
with a convolutional autoencoder and used it for anomaly detection. 
However, we consider that autoencoders 
that are only learned with unlabeled data 
do not fully leverage the expressive power of CNN.
Besides, recounting of abnormal events is yet to be considered,
while several approaches have been proposed for multimedia event recounting 
by localizing key evidence~\cite{Gan2015,Lai2014} or 
summarizing the evidence of detected events by text~\cite{Yu2012a}.


\section{Abnormal Event Detection and Recounting} 
We propose a method of detecting and recounting abnormal events.
As shown in Fig.~\ref{fig:overview} (a),
we learn generic knowledge about visual concepts 
in addition to learning environment-specific anomaly detectors.
Although most existing approaches use only environment-specific models,
they cannot extract semantic information and thus not sufficient to recount abnormal events.
Therefore, we learn the generic knowledge that is required for abnormal event recounting
by using large-scale supervised image datasets.
Since we learn the model with object, action, and attribute detection task that are highly related to abnormal event detection,
this generic model can be used to improve anomaly detection performance as shown in \cite{liu2015multi}.


First, multi-task Fast R-CNN is learned with large supervised datasets, 
which corresponds to the generic model that can be commonly used,
irrespective of the environment.
It is used to extract deep features 
(we call it a {\it semantic feature})
and visual concept classification scores from multiple local regions.
Second, anomaly detectors are learned on these features and scores
for each environment, 
which models the normal behavior of the target environment and
predict anomaly scores of test samples.
The anomaly detectors of features and classification scores 
are used for abnormal event detection and recounting, respectively.

Our abnormal event detection and recounting are performed 
using a combination of two learned models.
Figure~\ref{fig:overview} (b) outlines the four steps in the pipeline.
\begin{enumerate}
\setlength{\leftskip}{-7pt}
\setlength{\itemsep}{-4pt}
\setlength{\parsep}{0pt}
\item{\bf Detect object proposal:} Object proposals are detected for each frame by
geodesic object~\cite{Krahenbuhl2014} and moving object proposals~\cite{Fragkiadaki}.
\item{\bf Extract features:} Semantic features and classification scores are simultaneously extracted from all object 
proposals by the multi-task Fast R-CNN.
\item{\bf Classify normal/abnormal:} The anomaly scores of each proposal are computed by 
applying the anomaly detector to semantic features of the proposal.
The object proposals with anomaly scores above a threshold are determined as source regions of abnormal events.
\item{\bf Recount abnormal events:} Visual concepts of the three types (objects, actions, and attributes) 
of abnormal events are predicted from classification scores.
The anomaly scores of each predicted concept are computed by the anomaly detector for classification scores
to recount the evidence of anomaly detection. 
This phase is explained in more detail in Sec.~\ref{sec:method_recount}.
\end{enumerate}


\subsection{Learning of Generic Knowledge}
\label{sec:method_frcn}
We learn the generic knowledge about visual concepts to use it for
event recounting and to improve the performance 
of abnormal event detection. 
To exploit semantic information that is effective in these tasks,
we learn three types of concepts, i.e., {\it objects, actions, and attributes}, 
that are important to describe events. 
Since these concepts are jointly learned by multi-task learning, features
that are useful to detect any type of abnormality (abnormal objects, actions, or attributes)
can be extracted.
Our model is based on Fast R-CNN because it can efficiently predict categories 
and output features of multiple region-of-interests (RoIs) 
by sharing computation at convolutional layers.


{\bf Network architecture. }
Figure~\ref{fig:overview} (b) illustrates the architecture of the proposed multi-task Fast R-CNN
(shaded in red), which is the same as that for the Fast R-CNN except for the last classification layers.
It takes image and RoIs as inputs.
A whole image is first processed by convolutional layers
and its outputs are then processed by the RoI pooling layer and two fully-connected layers
to extract fixed length features from each RoI.
We used the feature at the last fully-connected layer (fc7 feature) 
as the semantic feature for learning abnormal event detector.
The features were fed into three classification layers, i.e.,
object, action, and attribute classification layers,
each of which consisted of fully-connected layers and activation.
A sigmoid was used for activation 
in attribute and action classification to 
optimize multi-label classification while softmax was used 
in object classification as in Girshick~\cite{Girshick2015}.
The bounding box regression was not used 
because it depends on the class to detect, which is not determined in abnormal event detection.
We used Alexnet~\cite{Krizhevsky2012} as the base network, which is commonly used as a feature 
extraction network and is computationally more efficient than that of VGG model~\cite{Simonyan2015}.

{\bf Training datasets.}
We used Microsoft COCO~\cite{Lin2014} training set to learn object and 
Visual Genome datasets~\cite{Krishna2016} to learn attributes and actions
because both datasets contain sufficiently large variations in objects with bounding box annotations.
Visual Genome was also used for the evaluation, as will be explained later in Sec.~\ref{sec:imeval}, 
and to seek for the fairness, the intersection of Visual Genome and COCO validation (COCO-val) set was excluded.
We used all 80 object categories in the COCO while 45 attributes and 25 actions that appeared 
the most frequently were selected from the Visual Genome dataset.
Our model only learned static image information using image datasets instead of video datasets because
motion information (e.g., optical flow) from the static camera was significantly different from 
that from the moving camera, and large datasets from the static camera with rich annotations were unavailable. 

{\bf Learning details.}
We used almost the same learning strategy and parameters as that for Fast R-CNN~\cite{Girshick2015}.
Here, we only describe differences from Fast R-CNN.
First, since we removed bounding box regression, our model was only trained with classification loss.
Second, our model was trained to predict multiple tasks, viz., object, action, and attribute detection.
A task was first randomly selected out of three tasks for each iteration, 
and a mini-batch was sampled from the dataset of the selected task following 
the same strategy as that for Fast R-CNN. 
The loss of each task was applied to its classification layer and shared layers.
Since multi-task model converged more slowly 
than the single-task model in \cite{Girshick2015},
we set the learning rate of SGD as 0.001 for first 200K iterations, 
and 0.0001 for the next 100K, 
which are larger numbers of iterations 
for each step of the learning rate than those for the single-task model.
All models are trained and tested with Chainer~\cite{tokui2015chainer}.

\subsection{Abnormal Event Recounting}
\label{sec:method_recount}
Abnormal event recounting is expected 
to predict concepts and also to provide evidence 
as to why the event was detected as an anomaly, 
which is not a simple classification task.
In the case in Fig.~\ref{fig:top} above, 
predicting category 
(object=`person', attribute=`young', and action=`bending')  
is not enough.
It is important to predict which concept is an anomaly 
({\it bending} is an anomaly)
to recount the evidence of abnormal events.
Therefore, as shown in Fig.~\ref{fig:top}, the proposed abnormal event recounting system predicts:
\begin{itemize}
\setlength{\itemsep}{-4pt}
\setlength{\parsep}{0pt}
\item{the categories of three types of visual concepts (object, action, and attribute) of the detected event}, and
\item{the anomaly scores for each concept to determine whether the evidence of detecting it as an anomaly.}
\end{itemize}

The approach to these predictions is straightforward.
We first predict categories 
by simply selecting the category with the highest classification score for each concept.
The anomaly score of each predicted category is then computed.
At training time, 
the distribution of classification scores under the target environment 
is modeled for each category by using kernel density estimation (KDE)
with a Gaussian kernel and a bandwidth calculated with 
Scott's rules~\cite{scott2015multivariate}.
At test time, 
the density at the predicted classification score 
is estimated by KDE for each predicted concept 
and the reciprocals of density are used as anomaly scores. 


\section{Experiments} 
\subsection{Datasets} 
\label{sec:abeval_set}
UCSD Ped2~\cite{Mahadevan2010} and Avenue~\cite{Lu2013} datasets were used 
to evaluate the performance of our method.
The UCSD pedestrian dataset is the standard benchmark for abnormal event detection,
where only pedestrians appear in normal events, while bikes, trucks, etc., appear in abnormal events. 
The UCSD dataset consists of two subsets, i.e., Ped1 and Ped2.
We selected Ped2 because Ped1 has a significantly lower frame resolution of 
158 $\times$ 240,
which would have made it difficult to capture objects in our framework based on object proposal+CNN. 
Since inexpensive higher resolution cameras have recently become commercially available,
we considered that this was not a critical drawback in our framework.
Avenue datasets~\cite{Lu2013} are challenging datasets 
that contain various types of abnormal events such as 
`throwing bag', `pushing bike', and `wrong direction'.
Since the pixel-level annotation in some complex events is subjective
(e.g., only the bag is annotated in a throwing bag event), 
we evaluated Avenue with only frame-level metrics.
In addition, while the Avenue dataset focuses on moving objects as abnormal events,
our focus included static objects.
Therefore, we evaluated the subset excluding five clips out of 22 clips that contained static 
but abnormal objects, viz., a red bag on the grass, and a person standing in front of a camera,
which are regarded as normal in the Avenue dataset.
We called this subset Avenue17, 
which we will describe in more detail in the supplemental material.
We used standard metrics in abnormal event detection, ROC curve, 
area under curve (AUC), and equal error rate (EER),
as was explained in Li et al.~\cite{Li2014} for both frame-level and pixel-level detection.

\begin{figure*}[t] 
\begin{center}
\footnotesize
\begin{tabular}{c} 
    \begin{minipage}{0.30\hsize} \begin{center}
        \includegraphics[width=1.00\linewidth]{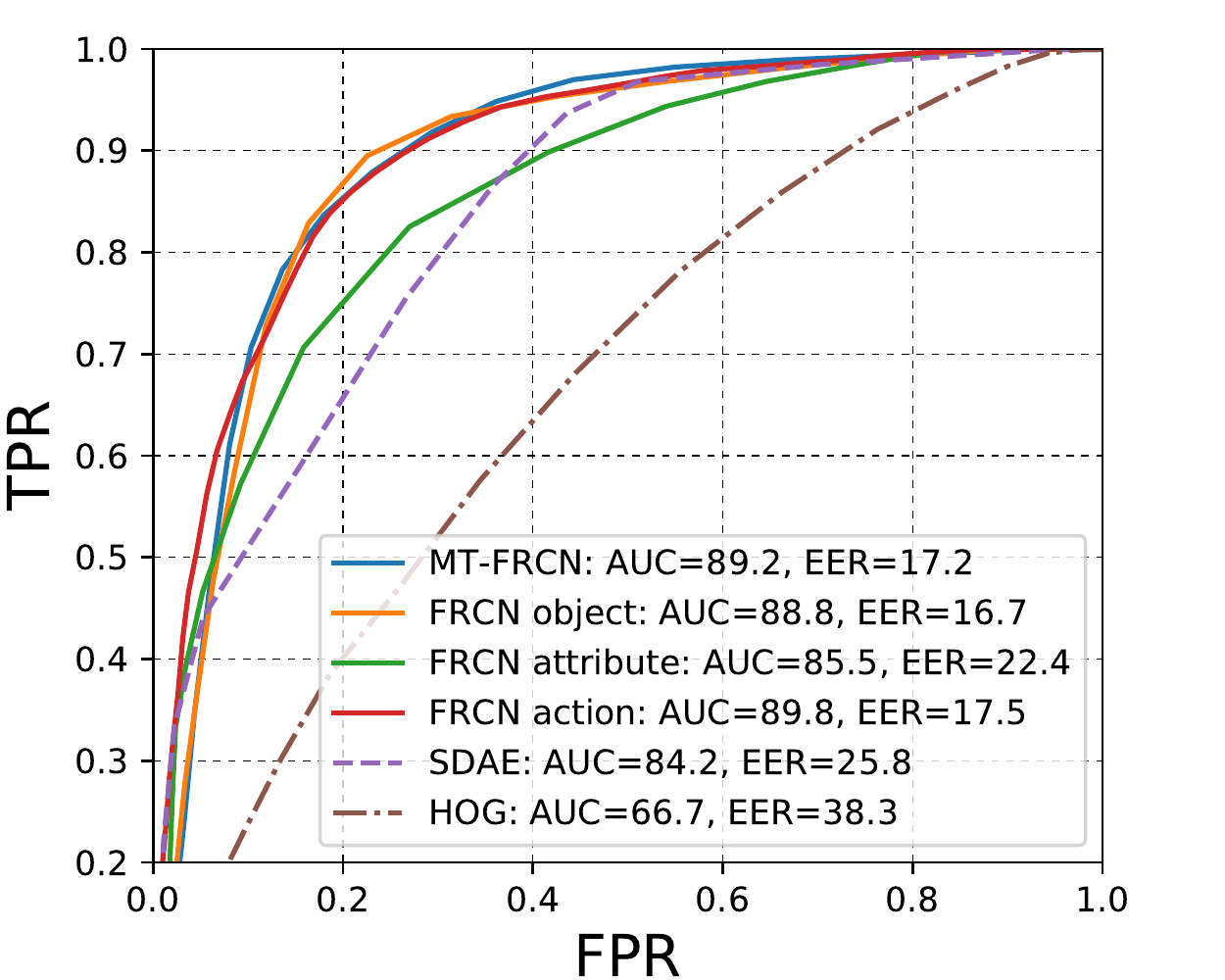} 
        (a) Avenue17
    \end{center} \end{minipage}
    \begin{minipage}{0.30\hsize} \begin{center}
        \includegraphics[width=1.00\linewidth]{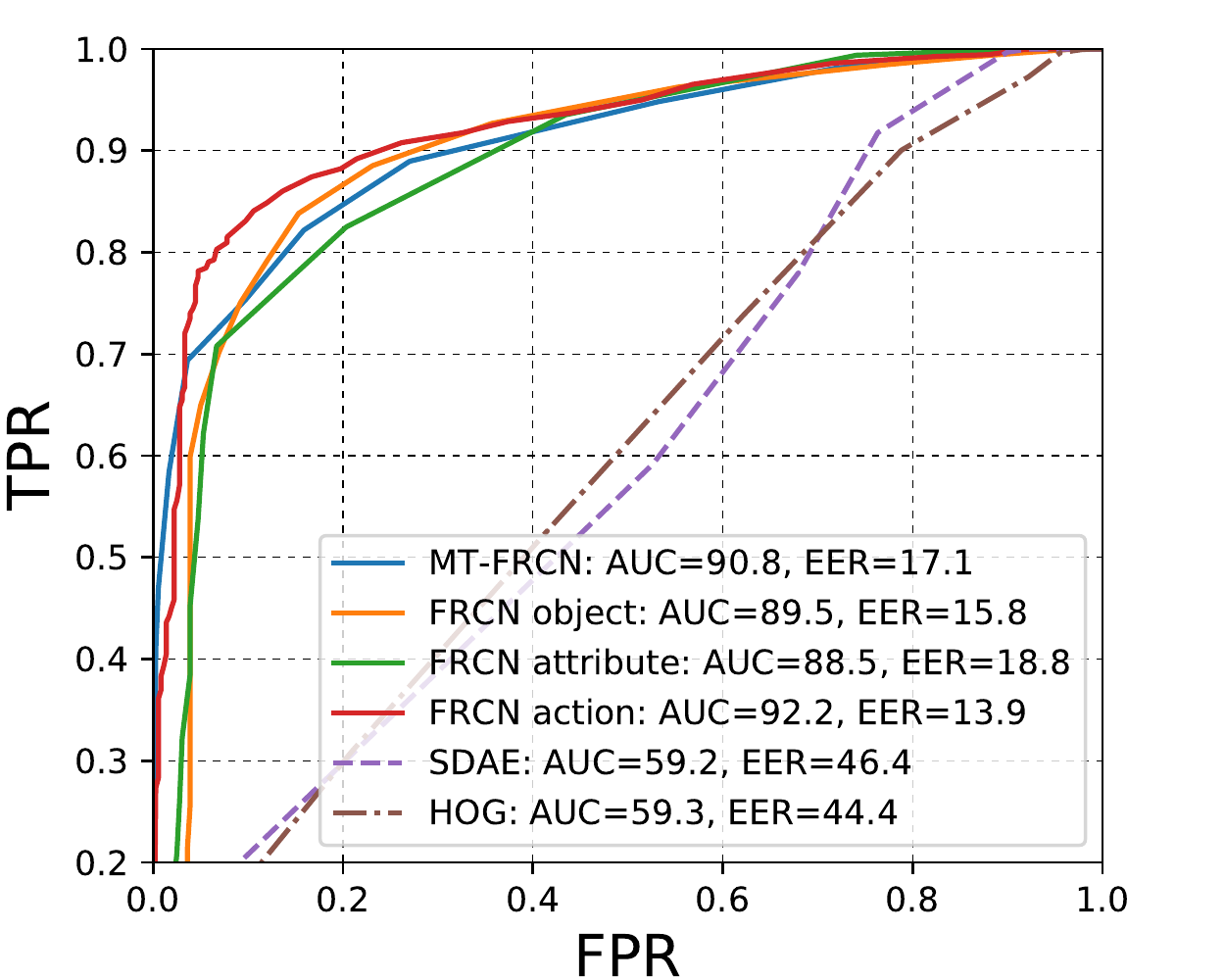} 
        (b) UCSD Ped2 (frame-level)
    \end{center} \end{minipage}
    \begin{minipage}{0.30\hsize} \begin{center}
        \includegraphics[width=1.00\linewidth]{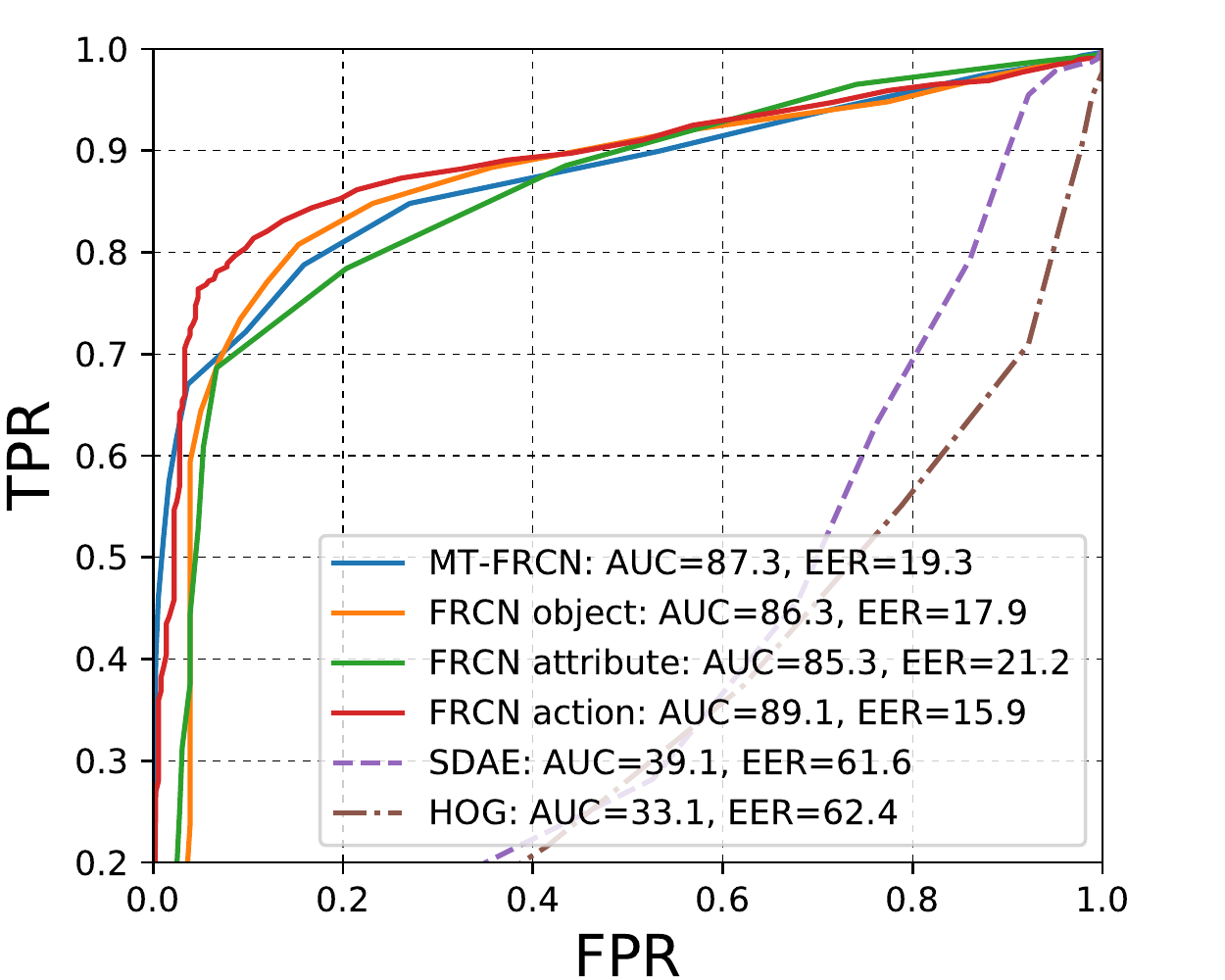} 
        (c) UCSD Ped2 (pixel-level)
    \end{center} \end{minipage}
\end{tabular}
\vspace{1mm}
\caption{Comparison of ROC curves between different appearance features
on standard benchmarks.}
\vspace{-5mm}
\label{fig:res_feat} \end{center} 
\end{figure*}
\subsection{Implementation details} 
\label{sec:imp}
{\bf Abnormal event detection procedure.}
Given the input video, we first detected object proposals from each frame 
using GOP~\cite{Krahenbuhl2014} and MOP~\cite{Fragkiadaki} as in \cite{Fragkiadaki}
(around 2500 proposals per frame).
The frame images and detected object proposals were input into Fast R-CNN
to obtain semantic features and classification scores for all proposals.
The semantic features were fed into the trained anomaly detector (described below)
to classify each proposal into normal or abnormal,
which computed an anomaly score for each proposal.
Object proposals with anomaly scores above the threshold were detected as abnormal events.
The threshold parameter was varied to plot the ROC curve in our evaluation.
Each detected event was finally processed for recounting, as was explained in Sec.~\ref{sec:method_recount}.

{\bf Anomaly detectors for semantic features.}
Given a training set extracted from training samples, 
anomaly detectors were learned to model `normal' behavior.
The anomaly detector took semantic features in testing as an input to
output an anomaly score.
Three anomaly detectors were used.
{\bf 1) Nearest neighbor-based method (NN): } An anomaly score was the distance between the test sample 
and its nearest training sample.
{\bf 2) One-class SVM (OC-SVM):} The anomaly score of test samples was the distance from the 
decision boundary of OC-SVM~\cite{Scholkopf2001} with RBF kernel. 
Since we did not have validation data, 
we did tests with several parameter combinations and used parameters that performed best ($\sigma$=0.001 and $\nu$=0.1).
{\bf 3) Kernel density estimation (KDE):}  Anomaly scores were computed as a reciprocal of density of
test samples estimated by KDE with a Gaussian kernel and a bandwidth calculated with 
Scott's rules~\cite{scott2015multivariate}.

To reduce computational cost, 
we separated frames into a 3$\times$4 grid with the same cell size, 
and learned the anomaly detectors for each location (12 detectors in total).
The coordinates of the bounding box center determined the cell that each object proposal belonged to. 
In addition,
features were compressed using product quantization (PQ)~\cite{Jegou2011} with a code length of 128 bits
in NN and features were reduced down to 16-dims using PCA in OC-SVM and KDE.

\subsection{Comparison of Appearance Features} 
\label{sec:abeval1}
We compare our framework using following different appearance features
to demonstrate the effectiveness of Fast R-CNN (FRCN) features in abnormal event detection: 
\begin{itemize}
\setlength{\leftskip}{-7pt}
\setlength{\itemsep}{-4pt}
\setlength{\parsep}{0pt}
\item {\bf HOG:} HOG~\cite{Dalal2005} extracted from a 32$\times$32 resized patch.
\item {\bf SDAE:} features of a stacked denoising autoencoder  
with the same architecture and training procedure as in \cite{Xu2015}.
\item {\bf FRCN objects, attributes, and actions:} The fc7 feature of single-task FRCN trained on one dataset.
\item {\bf MT-FRCN:} The fc7 feature of multi-task FRCN.
\end{itemize}
We used the same settings for other components 
including those for object proposal generation and anomaly detectors
to evaluate the effects of appearance features alone. 

{\bf ROC curves.}
Figure \ref{fig:res_feat} plots the ROC curves on Avenue17 and 
UCSD Ped2 datasets.
These experiments used NN as novelty detector. 
The curves indicate that FRCN features significantly outperformed HOG and SDAE in all benchmarks.
The main reason is FRCN features could discriminate different visual concepts
while HOG and SDAE features could not.
In the supplemental material,
the t-SNE map~\cite{VanDerMaaten2008} of feature space qualitatively justifies discriminability of each feature.
The FRCN action performs slightly better than the others
because the most challenging abnormal events in the benchmarks are related to actions.

\begin{figure}[t] 
\begin{center}
\footnotesize
\includegraphics[width=1.00\linewidth]{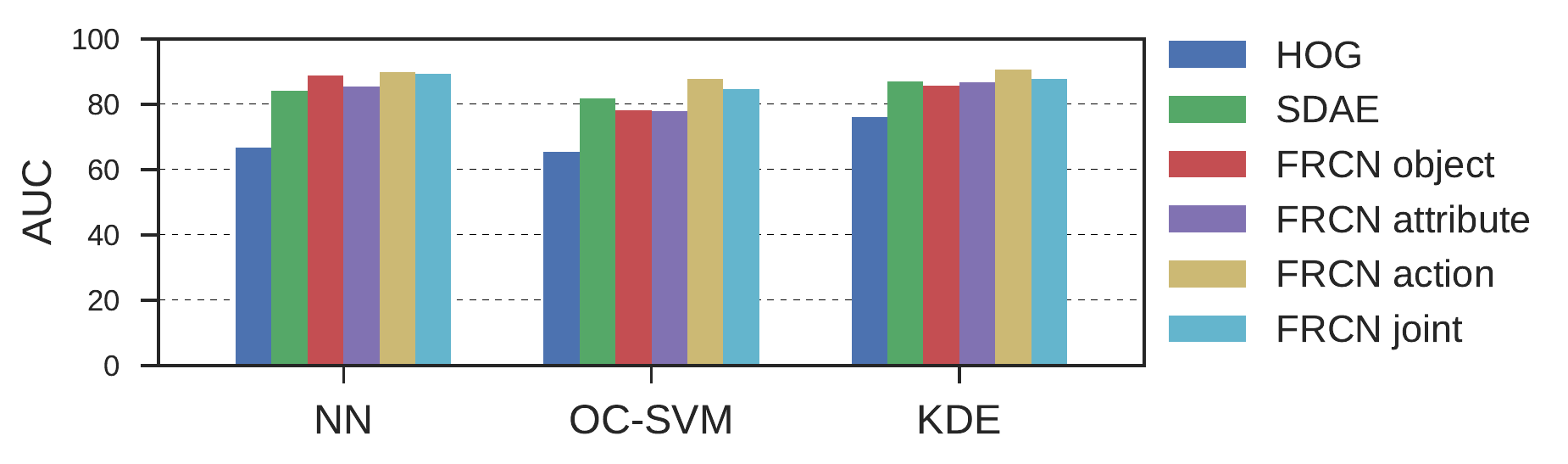} 
(a) Avenue17
\includegraphics[width=1.00\linewidth]{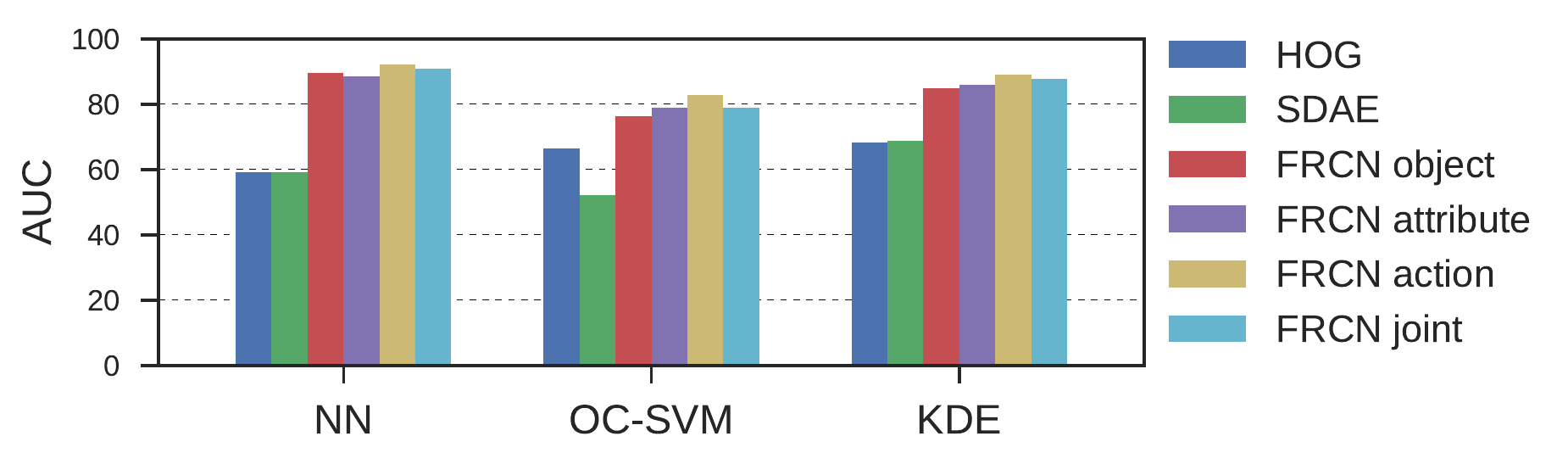} 
(b) UCSD Ped2 (frame-level)
\includegraphics[width=1.00\linewidth]{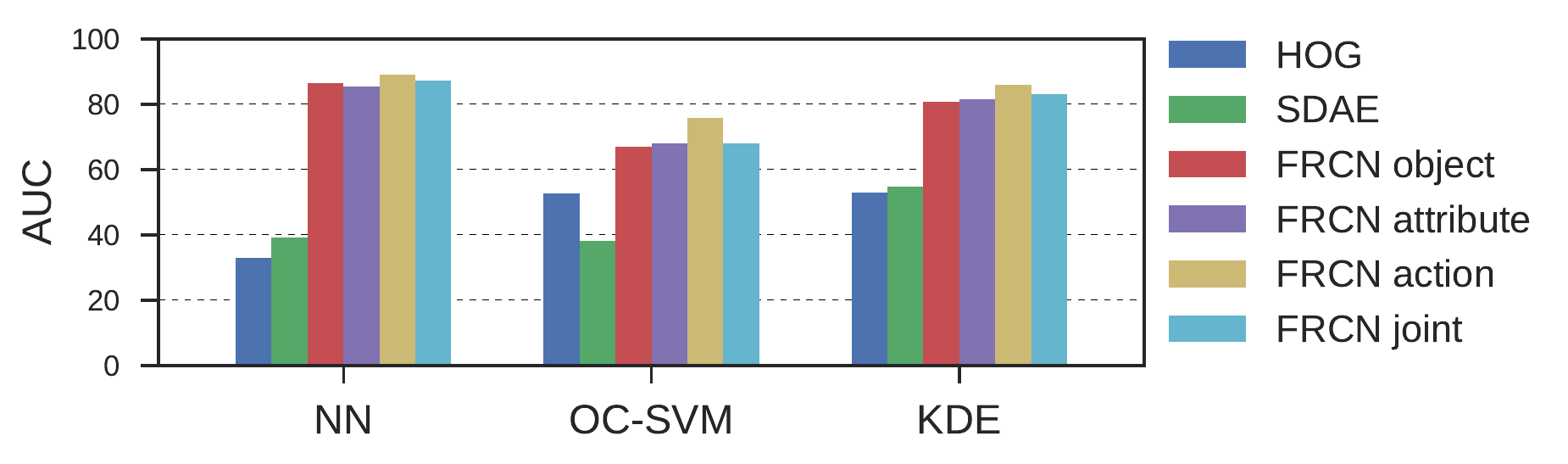} 
(c) UCSD Ped2 (pixel-level)
\caption{Comparison of AUC (\%) on standard benchmarks by varing anomaly detectors 
and appearance features.}
\vspace{-5mm}
\label{fig:res_det} \end{center} 
\end{figure}
{\bf Compatibility with different anomaly detectors.}
We measured performance using the three anomaly detectors explained in Sec.~\ref{sec:imp}
to clarify that FRCN features were compatible with various anomaly detectors,
Figure \ref{fig:res_det} compares AUC tested on varied anomaly detectors on Avenue17 and Ped2 datasets.
The results indicate that FRCN features always outperformed HOG and SDAE features 
and our performance was insensitive to anomaly detectors.
Since FRCN features were compatible with various anomaly detectors,
they can replace conventional appearance features 
in any framework for abnormal event detection.

\subsection{Comparison with State-of-the-art Methods} 
\label{sec:abeval2}
We compared our entire pipeline of abnormal event detection pipeline with state-of-the-art methods,
viz.,
local motion histogram (LMH)~\cite{Adam2008},
MPPCA~\cite{Kim2009},
social force model~\cite{Mehran2009}, 
MDT~\cite{Li2014},
AMDN~\cite{Xu2015},
and video parsing~\cite{Antic2011}
on the Ped2 dataset. 
We also made comparisons with Lu et al. (sparse 150 fps)~\cite{Lu2013}, and
Hasan et al. (Conv-AE)~\cite{Hasan2016} 
on the Avenue17 dataset.
We measured the performance of Avenue17 using the codes provided by the authors.

{\bf Results.}
Table~\ref{tab:sota} summarize the AUC/EER on Avenue17 and UCSD Ped2 datasets, 
which demonstrates our framework outperformed all other methods on all benchmarks.
Especially, AUC of Ped2 was 89.2\%, which significantly outperformed the state-of-the-art method 
(66.5\% in \cite{Li2014}).
Since our method was based on object proposals and captured object-level semantics by using FRCN features, 
we accurately localized abnormal objects.
Moreover, the Avenue17 dataset contained objects and actions  
that were not included in Fast R-CNN's training data (e.g., white paper and throwing bag). 
This indicated that FRCN features generalized the detection of untrained categories.
Note that our method performed best without using any motion features
while others used motion features based on optical flows.
Learning motion features with two-stream CNN~\cite{Feichtenhofer2016} or 3D-CNN~\cite{Tran2015} 
remains to be undertaken in future work.
Also, our performance on Ped1 is much worse than state-of-the-art (69.9/35.9 in AUC/EER)
because of the low resolution issue as stated above, which should be solved in the future.

\begin{table} 
\begin{center}
\small
\begin{tabular}{@{}l c c c@{}}
\toprule
Method                               & Avenue17 & Ped2 (frame)& Ped2 (pixel)\\
\midrule
LMH~\cite{Adam2008}                  & -         & 69.3/30   & 15.9/77.6 \\
MPPCA~\cite{Kim2009}                 & -         & 69.3/30   & 22.2/77.6 \\
Social force~\cite{Mehran2009}       & -         & 55.6/42   & 21.7/72.4 \\
MDT~\cite{Li2014}                    & -         & 82.9/27.9 & 66.5/29.9 \\
AMDN~\cite{Xu2015}                   & -         & 90.8/17   & -       \\
Video parsing~\cite{Antic2011}       & -       & 92/-      & -       \\
Sparse 150fps~\cite{Lu2013}          & 80.3/27.5 & -         & -         \\
Conv-AE~\cite{Hasan2016}             & 76.9/34.0 & 90.0/21.7 & -       \\
\midrule
FRCN object                          & 88.8/{\bf 16.7} & 89.5/15.8 & 86.3/19.3 \\
FRCN attribute                       & 86.7/22.7 & 88.5/18.8 & 85.3/21.2 \\
FRCN action                          & {\bf 89.8}/17.5 & {\bf 92.2}/{\bf 13.9} & {\bf 89.1}/{\bf 15.9} \\
MT-FRCN                              & 89.2/17.2 & 90.8/17.1 & 87.3/19.4 \\
\bottomrule
\end{tabular}
\end{center}
\vspace{-2mm}
\caption{Abnormal event detection accuracy in AUC/EER (\%).}
\label{tab:sota}
\vspace{-2mm}
\end{table}
\begin{figure*}[t] 
\begin{center}
\includegraphics[width=1.00\linewidth]{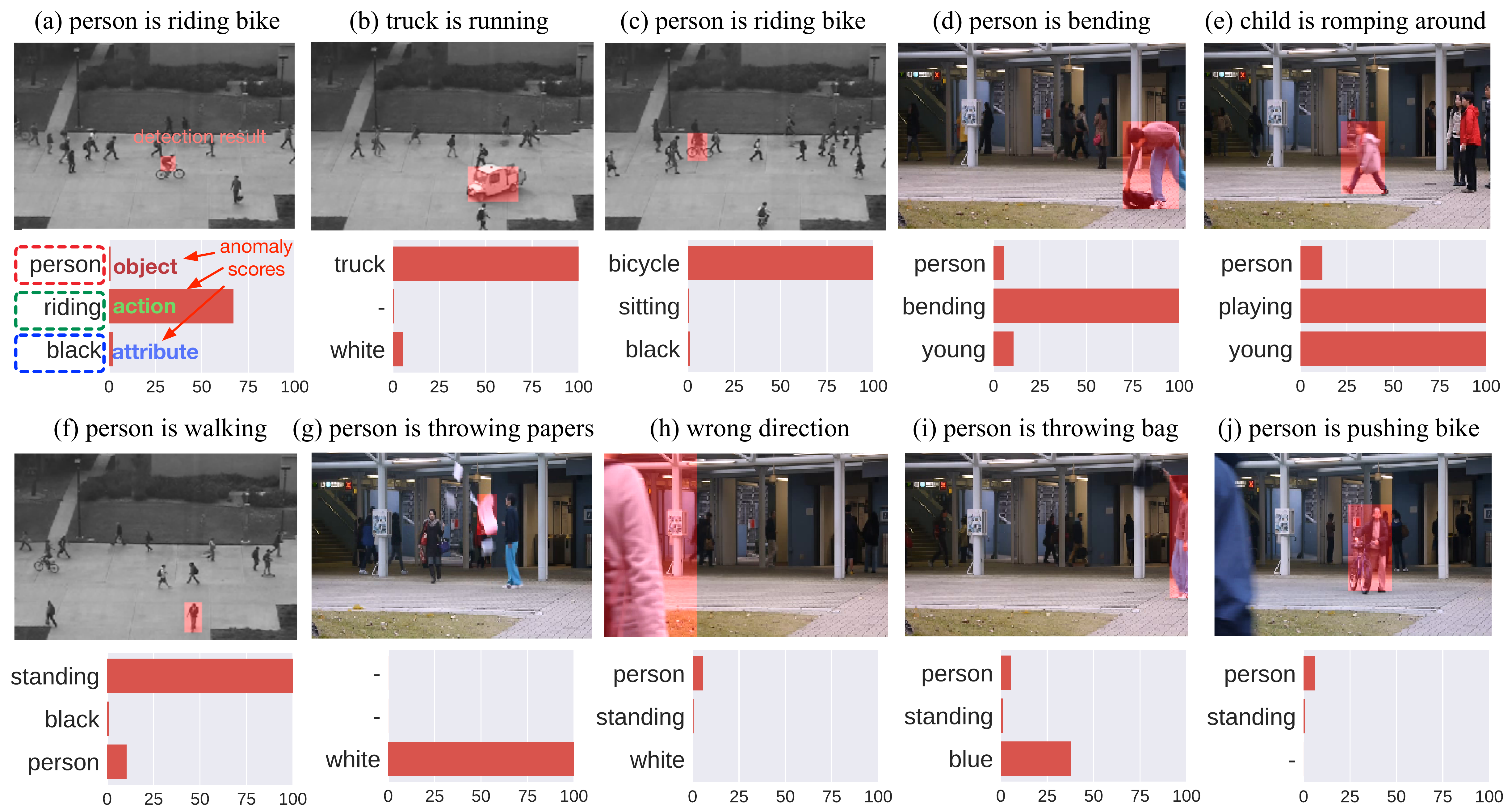}
\caption{Examples of recounting. 
Each example shows the recounting result of one detected event shaded in red.
Its predicted categories and anomaly scores of each category (red bars) are indicated as recounting results. 
`-' corresponds to where classification scores of all categories are under 0.1.}

\vspace{-2mm}
\label{fig:recount} 
\end{center} 
\end{figure*}
\subsection{Qualitative Evaluation of Evidence Recounting} 
\label{sec:abeval3}
Figure~\ref{fig:recount} has examples of recounting results obtained with our framework
where the predicted categories and anomaly scores of each category (red bars) have been presented.
Figures~\ref{fig:recount} (a)--(e) present successfully recounted results.
Our method could predict abnormal concepts such as `riding', `truck', and `bending'
while assigning lower anomaly scores to normal concepts such as `person' and `black'.
The anomaly score of `young' in (e) is much higher than that in (d) because a high 
{\it classification} score for `young' was assigned to the child in (e), which is rare.
Figures~\ref{fig:recount} (f)--(j) reveal the limitations of our approach.
The event in (f) is a false positive detection. Since we only used appearance information,
a person walking in a different direction from the other people is predicted as standing.
The events in (g) and (h), viz., scattered papers and the person in the wrong direction, 
could not be recounted with our approach because they were outside the knowledge we learned.
Nevertheless, the results provided some clues to understanding events; the event (g) is something `white'
and the anomaly in the event (h) is not due to basic visual concepts.
The events in (i) and (j) that correspond to `throwing a bag' and `pushing a bicycle' 
include the interaction of objects, 
which could be captured with our approach.
Since large datasets for object interactions is available~\cite{Chao2016,Krishna2016,Le2014}, 
our framework could be extended to learn such knowledge, 
and this could be another direction for future work.


\section{Evaluation with Artificial Datasets} 
\label{sec:imeval}
\subsection{Settings}
The current benchmark in abnormal event detection has three main drawbacks when evaluating our methods.
1) The dataset size is too small and variations in abnormalities are limited because
collecting data on abnormal events is difficult due to privacy issues in surveillance videos
and the scarcity of abnormal events.
2) The definition of abnormal is subjective because it depends on applications.
3) Since ground truth labels on the categories of abnormal events are not annotated,
it is difficult to evaluate recounting.

The experiments described in this subsection were designed to evaluate the performance 
of {\it unseen (novel) visual concept detection},
i.e., detect basic visual concepts 
that did not appear in the training data,
which represent an important portion of abnormal event detection.
Although most events in the UCSD pedestrian dataset belong to this category, 
the variations in concepts are limited (e.g., person, bikes, and trucks).
We artificially generated the dataset of unseen visual 
concept detection with large variations based on image dataset.
Its evaluation scheme was more objective than 
abnormal event detection benchmarks.

{\bf Task Settings.}
This task was evaluated on the dataset with the 
bounding box annotations of objects, actions, or attributes.
The $n_{seen}$ categories were selected from all $n$ annotated categories and
the dataset was split into training and test sets
so that training set only contained $n_{seen}$ categories. 
The main objective of this task was to find $n_{unseen}=n-n_{seen}$ categories from the test set.
In other words, we detected unseen categories that did not appear in the training set,
which had similar settings to abnormal event detection benchmarks.
We specifically propose two tasks to evaluate our method.
{\bf Task 1 (Sec.~\ref{sec:imeval1}):} Detect objects that have annotations of unseen categories
using our abnormal event detection framework (anomaly detector + fc7 features).
{\bf Task 2 (Sec.~\ref{sec:imeval2}):} Detect and classify unseen objects 
with our method of abnormal event recounting 
(kernel density estimation + classification scores).

\begin{table*} 
\begin{center}
\begin{tabular}{@{}l c c c c c c c c c c c@{}}
\toprule
 & \multicolumn{5}{c}{{\bf training data for feature learning}}  & \multicolumn{3}{c}{} \\ 
 & \multicolumn{2}{c}{unlabeled} & \multicolumn{3}{c}{labeled}  & \multicolumn{3}{c}{{\bf mAP (COCO)}} & \multicolumn{3}{c}{{\bf mAP (PASCAL)}} \\ 
 \cmidrule(lr){2-3}\cmidrule(lr){4-6}\cmidrule(lr){7-9}\cmidrule(l){10-12}
{\bf Feature} & generic & normal      & object & attribute & action & object   & attribute & action   & object   & attribute & action   \\
\midrule
HOG                  &        &        &        &        &        & 8.3      & 5.4      & 16.8       & 34.6     &  2.4      & 42.3     \\\midrule
SDAE                 & \cmark &        &        &        &        & 8.1      & 5.6      & 17.5       & 34.8     &  1.0      & 42.7    \\
SDAE                 &        & \cmark &        &        &        & 10.9     & 5.2      & 17.9       & 34.6     &  1.6      & 43.2     \\\midrule
FRCN object      &        &        & \cmark &        &        &{\bf 20.6}& 13.6     & 20.5       &{\bf 62.0}&  1.9      & 47.9     \\
FRCN attribute   &        &        &        & \cmark &        & 12.1     &{\bf 18.3}& 20.9       & 48.3     &{\bf 2.8}  & 47.1     \\
FRCN action      &        &        &        &        & \cmark & 12.4     & 9.1      & {\bf 30.2} & 44.5     &  1.8      &{\bf 50.8}\\
MT-FRCN          &        &        & \cmark & \cmark & \cmark & 16.9     & 14.5     & 29.6       & 57.7     &  2.2      & 50.2     \\
\bottomrule
\end{tabular}
\end{center}
\vspace{-2mm}
\caption{Performance of unseen visual concept detection on artificially created dataset based on COCO and PASCAL datasets.
Training data used for feature learning are also indicated by check marks.}
\label{tab:unseen} 
\vspace{-2mm}
\end{table*}
\subsection{Evaluation of Unseen Concept Detection}
\label{sec:imeval1}
We evaluated this task on the datasets based on COCO and PASCAL datasets.
We used the COCO-val set for objects, and the intersection of COCO-val and Visual Genome for actions and attributes.
We used the same categories that were used to train Fast R-CNN.
As for PASCAL dataset, official PASCAL VOC 2012 datasets were used for object and action detection, 
while the a-PASCAL dataset~\cite{Farhadi2012} was used for attribute detection.
Each dataset was split into training and test sets in the following procedure: 
1) randomly select unseen categories ($n_{unseen}$ is set to be around $n$/4), 
2) assign images with unseen category objects to training sets, 
3) assign randomly sampled images to training sets as distractors 
(so that {\it\#} of test images equal to {\it\#} of training images), and
4) assign remaining images to test sets.
We repeated this to create five sets of training--test pairs for each dataset.

We used the same method of detection as that in the experiments in Sec.~\ref{sec:abeval1}; 
unseen categories were detected as regions with high anomaly scores computed by
a nearest neighbor-based anomaly detector trained for each training set.
We used the ground truth bounding boxes as input RoIs instead of 
using object proposals because some proposals contained unannotated but unseen categories, 
which made it difficult to evaluate our framework.
To evaluate performance, 
detection results are ranked by the anomaly score 
and average precision (AP) was calculated similarly to PASCAL detection
(objects with the annotations of unseen categories are positive in our evaluation).
The final performance values were computed as mean average precision (mAP) 
over the five sets of unseen categories.

Table~\ref{tab:unseen} summarizes the mAP of our framework with different appearance features.
The training data to train each feature have also been listed as check marks.
We trained two SDAE: a {\it generic} model trained on the dataset used in Fast R-CNN learning, 
and a {\it specific} model trained on the training data of each set (that only contained `seen' categories).
The results demonstrated that Fast R-CNN significantly outperformed HOG and SDAE, which
indicated that unseen visual concept detection is a difficult task without learning with labeled data.
The single-task Fast R-CNN trained on the same task as the evaluation task 
performed best in all tasks while the proposed multi-task Fast 
R-CNN gained the second highest mAP in all tasks, which was significantly better 
than models trained on different tasks.
Since the types of abnormal concepts to be detected were not fixed in practice,
multi-task Fast R-CNN is an excellent choice for abnormal event detection.

\subsection{Evaluation of Unseen Concept Recounting}
\label{sec:imeval2}
We quantitatively evaluated our recounting method by using 
the COCO-based unseen concept detection dataset in Sec.~\ref{sec:imeval1}.
For each candidate region of test sample, 
our framework outputs the classification scores 
and anomaly scores computed by KDE learned from the train set.
The performance values were computed as AUC of TPR versus FPR. 
For a certain threshold of anomaly scores, unseen categories were predicted for each region, i.e., 
categories with the anomaly scores above the threshold and classification scores above 0.1.
Unlike the experiments described in Sec.~\ref{sec:abeval3}, 
multiple categories were sometimes predicted for each concept in this evaluation.
An object was true positive if 1) ground truth unseen categories were annotated (it was positive), and
2) the predicted unseen categories agreed with the ground truth.
An object was false positive if 1) ground truth unseen categories were not annotated (it was negative), and
2) any category was predicted as being unseen.
The threshold was varied to compute AUC.
We compared our method with HOG and SDAE features combined with a linear SVM classifier.
The SVM classification scores were used as the input for the anomaly detector in these methods.
SVMs were trained on the COCO-training set that was used in Fast R-CNN training.

Table~\ref{tab:res_recount} compares AUC on the COCO-based unseen concept detection datasets.
We can see that multi-task Fast R-CNN outperformed best
with all types of concepts 
while HOG and SDAE could hardly recount unseen concepts.
This demonstrates that deeply learned generic knowledge 
is essential for concept-level recounting of abnormal events.

\begin{table} 
\begin{center}
\small
\begin{tabular}{@{}l c c c@{}}
\toprule
Method & object & attribute & action \\
\midrule
HOG+SVM           & 2.1 & 1.2 & 0.8  \\
SDAE+SVM          & 2.3 & 1.2 & 1.3  \\
MT-FRCN+SVM        & 24.7 & 13.4 & 16.2  \\
MT-FRCN output     & {\bf 26.8} & {\bf 15.4} & {\bf 16.5}   \\
\bottomrule
\end{tabular}
\end{center}
\caption{AUC of recounting on artificially created abnormal concept detection dataset based on COCO. }
\label{tab:res_recount}
 \vspace{-3mm}
\end{table}


\section{Conclusion} 
We addressed the problem of joint abnormal event detection and recounting.
To solve this problem, we incorporate the learning of
generic knowledge, which is required for recounting, and environment-specific knowledge, 
which is required for anomaly detection, into a unified framework.
Multi-task Fast R-CNN is first trained on richly annotated image datasets 
to learn generic knowledge about visual concepts.
Anomaly detectors are then trained on the outputs of this model to
learn environment-specific knowledge.
Our experiments demonstrated the effectiveness of our method for abnormal event detection and recounting
by improving the state-of-the-art performance on challenging benchmarks 
and providing successful examples of recounting.

Although this paper investigated basic concepts such as actions,
our approach could be extended further to complex concepts such as object interactions.
This work is the first step in abnormal event detection using generic knowledge of visual concepts and 
sheds light on future directions for such higher-level abnormal event detection.

{\bf Acknowledgements:}
We thank Cewu Lu, Allison Del Giorno, and Mahmudul Hasan 
for sharing their code and data.
This work was supported by JST CREST JPMJCR1686 
and JSPS KAKENHI 17J08378.

\clearpage
{\small
\bibliographystyle{ieee}
\bibliography{AbnormalEventDetection,add}
}

\end{document}